\documentclass[sigconf]{acmart}
\usepackage{algorithm}
\usepackage{algorithmic}

\AtBeginDocument{%
  \providecommand\BibTeX{{%
    \normalfont B\kern-0.5em{\scshape i\kern-0.25em b}\kern-0.8em\TeX}}}

\setcopyright{rightsretained}
\copyrightyear{2024}
\acmYear{2024}
\acmDOI{10.1145/3689096.3689459}

\acmConference[VLM4Bio '24]{Proceedings of the First International Workshop on Vision-Language Models for Biomedical Applications}{October 28-November 1, 2024}{Melbourne, VIC, Australia}
\acmBooktitle{Proceedings of the First International Workshop on Vision-Language Models for Biomedical Applications (VLM4Bio '24), October 28-November 1, 2024, Melbourne, VIC, Australia}
\acmISBN{979-8-4007-1207-4/24/10}

\acmSubmissionID{vlm0903}

\begin{document}

\title{A Causal Approach to Mitigate Modality Preference Bias in Medical Visual Question Answering}

\renewcommand{\shorttitle}{MedCFVQA}


\author{Shuchang Ye}
\orcid{0009-0006-1935-1953}
\affiliation{%
  \institution{The University of Sydney}
  \city{Sydney}
  \country{Australia}}
\email{shuchang.ye@sydney.edu.au}

\author{Usman Naseem}
\affiliation{%
  \institution{Macquarie University}
  \city{Sydney}
  \country{Australia}}
\email{usman.naseem@sydney.edu.au}

\author{Mingyuan Meng}
\affiliation{%
  \institution{The University of Sydney}
  \city{Sydney}
  \country{Australia}}
\email{mmen2292@uni.sydney.edu.au}

\author{Dagan Feng}
\affiliation{%
  \institution{The University of Sydney}
  \city{Sydney}
  \country{Australia}}
\email{dagan.feng@sydney.edu.au}

\author{Jinman Kim}
\affiliation{%
  \institution{The University of Sydney}
  \city{Sydney}
  \country{Australia}}
\email{jinman.kim@sydney.edu.au}

\renewcommand{\shortauthors}{Ye, et al.}

\begin{abstract}
  Medical Visual Question Answering (MedVQA) is crucial for enhancing the efficiency of clinical diagnosis by providing accurate and timely responses to clinicians' inquiries regarding medical images. Existing MedVQA models suffered from modality preference bias, where predictions are heavily dominated by one modality while overlooking the other (in MedVQA, usually questions dominate the answer but images are overlooked), thereby failing to learn multimodal knowledge. To overcome the modality preference bias, we proposed a Medical CounterFactual VQA (MedCFVQA) model, which trains with bias and leverages causal graphs to eliminate the modality preference bias during inference. Existing MedVQA datasets exhibit substantial prior dependencies between questions and answers, which results in acceptable performance even if the model significantly suffers from the modality preference bias. To address this issue, we reconstructed new datasets by leveraging existing MedVQA datasets and Changed their P3rior dependencies (CP) between questions and their answers in the training and test set.   Extensive experiments demonstrate that MedCFVQA significantly outperforms its non-causal counterpart on both SLAKE, RadVQA and SLAKE-CP, RadVQA-CP datasets.
\end{abstract}



\begin{teaserfigure}
\centering
  \includegraphics[width=0.9\textwidth]{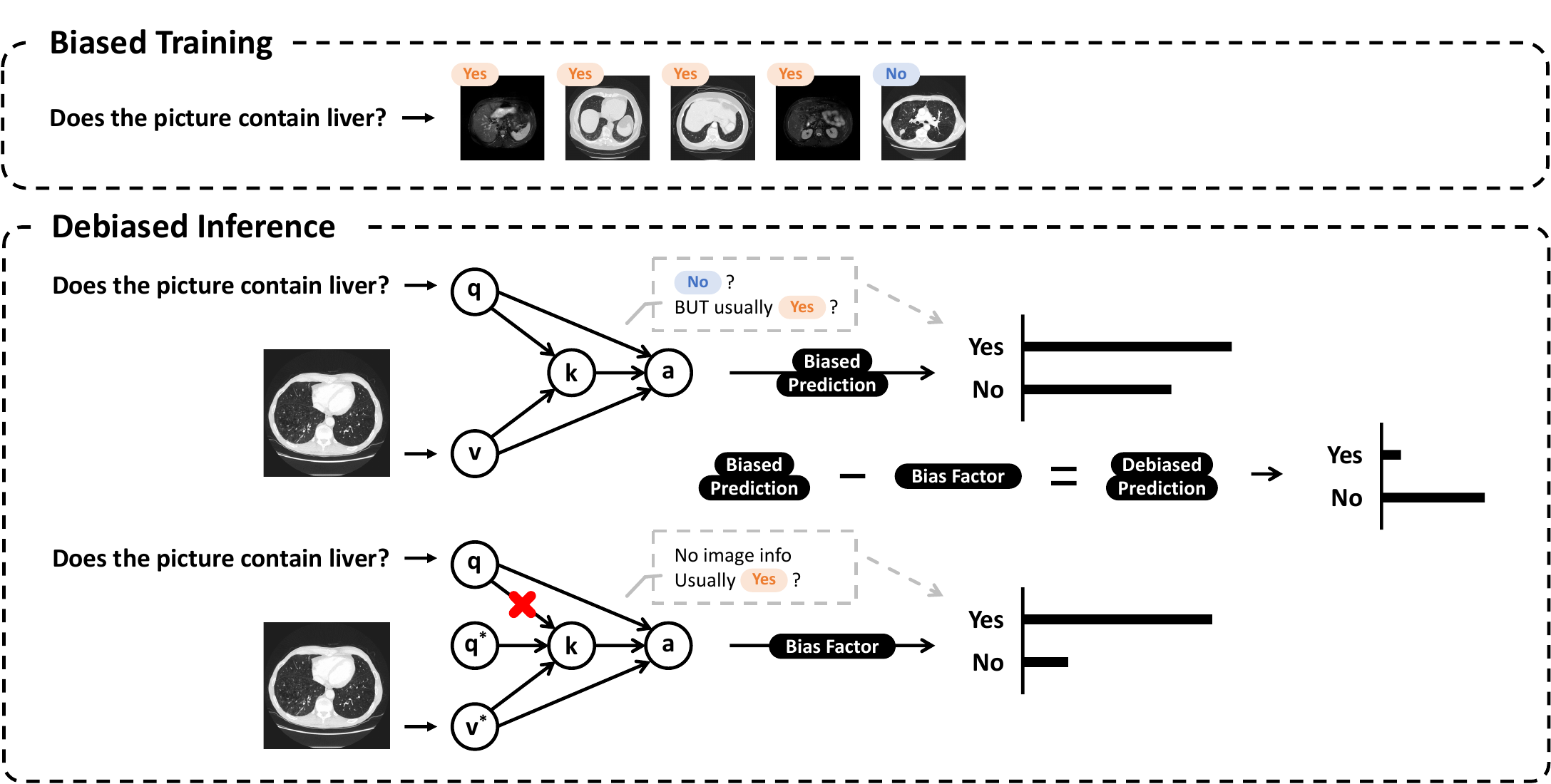}
  \caption{Overview of our MedCFVQA model. The model initially suffers from modality preference bias during training. During inference, the model captures and subtracts this bias. An asterisk (*) indicates missing information.}
  \Description{MedCFVQA}
  \label{fig:medcfvqa}
\end{teaserfigure}


\maketitle

\section{Introduction}
Medical Visual Question Answering (MedVQA) is essential for achieving efficient diagnosis in modern healthcare~\cite{survey2023}, with applications spanning radiology~\cite{vqa_in_rad}, pathology~\cite{vqa_in_path}, dermatology~\cite{vqa_in_der}, and chat-based medical consultations~\cite{vqa_in_chat}. Despite improvements in the accuracy of MedVQA methods, recent studies~\cite{vqa-cp, rubi, cfvqa} have revealed that the answers often do not derive from a true causal relationship involving the images and questions, but instead rely on superficial correlations between the questions and answers (Figure~\ref{fig:medcfvqa}), which is called modality preference bias in our study. In the SLAKE~\cite{slake} dataset, the question "Are there abnormalities in this image?" shows a highly imbalanced response distribution, with "yes" dominating across all splits: 59:1 in the training set, 9:1 in the validation set, and 12:0 in the test set. This bias leads models to predict "yes" regardless of the image, resulting in misleadingly high accuracy—up to $90\%$ in validation and $100\%$ in testing. This modality preference bias, where the questions are preferable than input images, renders the model ineffective when the training data distribution differs from real-world clinical scenarios.

Neural network debiasing methods can be categorized into dataset, training, and inference approaches. Dataset debiasing, through undersampling~\cite{cnn_imbalance, overparameterization} and oversampling~\cite{oversample}, often results in underfitting or overfitting due to the loss or repetition of samples. Training debiasing methods, such as up-weighting minority cases~\cite{stochastic_gradients} and generating adversarial samples~\cite{biasadv}, are limited in effectiveness, particularly in the medical field where data scarcity hinders robust adversarial generation. Inference debiasing~\cite{rubi,cfvqa}, which addresses modality preference bias during the prediction phase, leverages the full dataset for training and avoids introducing new errors, making it particularly suitable for clinical applications where data limitations are prevalent. However, inference debiasing methods remain underexplored in the medical domain.

Eliminating modality preference bias during inference is based on counterfactual inference of causal reasoning theory~\cite{cfvqa}. As illustrated in Figure~\ref{fig:medcfvqa}, we propose a novel Medical CounterFactual VQA (MedCFVQA) to address the modality preference bias by subtracting the bias during inference. In MedVQA, the causality relation can be represented as follows: a question ($q$) and an image ($v$) cause multimodal knowledge ($k$), which then determines the answer ($a$). Modality preference bias is represented by the direct causal effect~\cite{direct_effect} $v/q \rightarrow a$. To address this, we employ counterfactual reasoning by considering the counterfactual scenario where $q$ is present but $v$ and $k$ are absent, thus estimating what a would be under these conditions. This counterfactual approach allows us to isolate and subtract the spurious correlation from the biased causality relation (see Figure~\ref{fig:medcfvqa}). When facing the question, ``\textit{Does the picture contain a liver?}" and an image lacking a liver, conventional VQA models might erroneously answer ``\textit{yes}" due to the spurious correlation having a stronger influence on the answer than multimodal knowledge. However, by applying counterfactual inference to compute and remove this bias, the system can correct for the modality preference bias, yielding an answer that accurately reflects the true multimodal knowledge based on the image content.

The modality preference bias behavior is underestimated because the train-test splits of existing MedVQA datasets when strong priors are present. This leads models to perform well on test sets by memorizing biases from the training data rather than understanding the multimodal knowledge. To address this, we create the SLAKE-CP and RadVQA-CP datasets, which re-distribute the train and test splits of SLAKE~\cite{slake} and RadVQA~\cite{radvqa}. These new splits ensure that the distribution of answers per question differs between the training and test sets. For instance, for the question ``\textit{Does the rectum appear in the image?}" the answer ``\textit{yes}" appears exclusively in either the training or testing set, but not both.  

Our contributions can be summarized as follows:
\begin{itemize}
    \item We propose a novel medical counterfactual VQA model (MedCFVQA), utilizing causal reasoning to identify and subtract modality bias during inference. 
    \item We apply greedy re-splitting algorithm~\cite{vqa-cp} to MedVQA datasets, SLAKE and RadVQA, and created the SLAKE-CP and RadVQA-CP datasets for modality preference bias benchmarking.
    \item Extensive experiments shows that MedCFVQA outperforms conventional VQAs on both SLAKE, RadVQA and SLAKE-CP, RadVQA-CP datasets.
\end{itemize}

\begin{figure*}
  \includegraphics[width=0.8\textwidth]{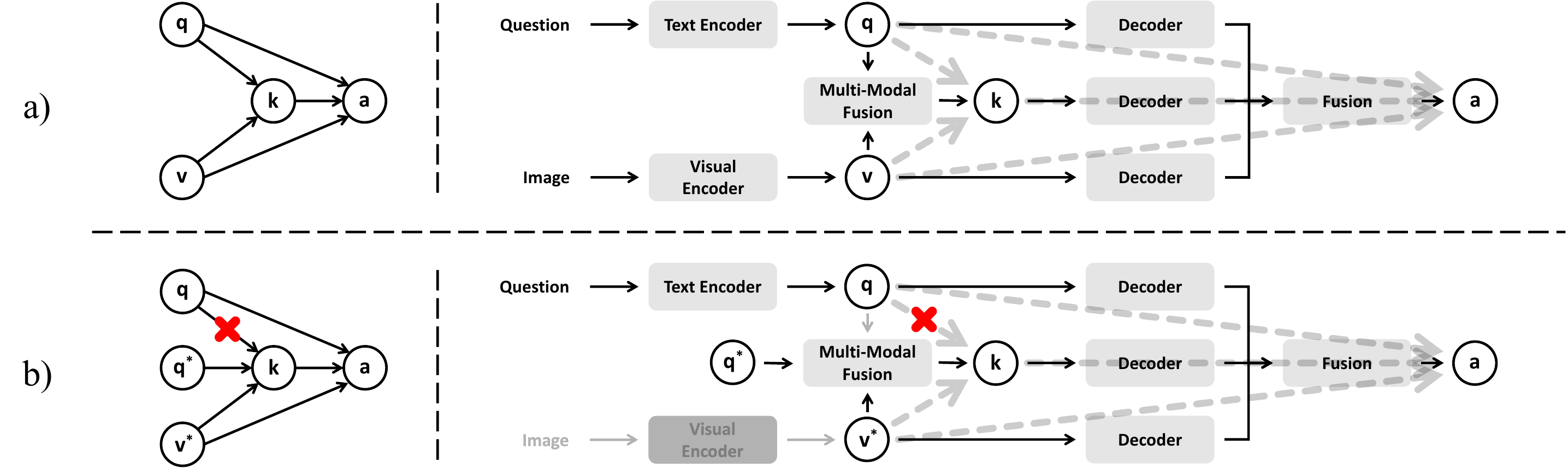}
  \caption{Integration of causal graph and neural network: The left panel illustrates the causal graph, while the right panel depicts the neural network architecture. An asterisk (*) indicates that the item is counterfactual. $x \rightarrow y$ represents the direct causal effect of $x$ on $y$, while $x \rightarrow m \rightarrow y$ represents the indirect causal effect of $x$ on $y$.}
  \Description{Causal Graph Meets Neural Network}
  \label{fig:meet}
\end{figure*}

\section{Method}

\subsection{Greedy Re-splitting}

Existing MedVQA datasets often demonstrate substantial prior dependencies between questions and answers, which can mislead models into achieving deceptively high performance. To overcome the prior and provide a effective benchmark of modality preference bias, we constructed the SLAKE-CP and RadVQA-CP datasets by re-splitting the original SLAKE~\cite{slake} and RadVQA~\cite{radvqa} datasets to alter the distribution of answers per question between the training and testing sets. This process ensures that the training set fully covers all words in the test set, while the test set maximizes the coverage of concepts present in the training set. The applied Greedy Re-splitting~\cite{vqa-cp} is implemented as shown in Algorithm~\ref{alg:gr}.

\begin{algorithm}[t]
\caption{Greedy Re-splitting}
\label{alg:gr}
\begin{algorithmic}[1] 
\STATE G $\leftarrow$ Group samples with same <question, answer> pair
\STATE C $\leftarrow$ set(words in $G$) 
\STATE train, test $\leftarrow$ [], []
\STATE $C_{train}$, $C_{test}$ $\leftarrow$ set(), set()
\STATE $R$ $\leftarrow$ copy($G$)
\WHILE{Not sufficient samples in test set}
\STATE Grouptest $\leftarrow$ group in $R$ 
\STATE test.append(Grouptest)
\STATE $C_{test}$.append(concepts(Grouptest))
\STATE $R$ -= Grouptest
\STATE maxGrouptrain $\leftarrow$ group in $R$ with $max(C_{group} \cap C_{test}-C_{train})$
\STATE train.append(maxGrouptrain)
\STATE $C_{test}$.append(concepts(maxGrouptrain)) 
\STATE $R$ -= maxGrouptrain
\ENDWHILE
\STATE train $\leftarrow$ $R$
\end{algorithmic}
\end{algorithm}

\begin{figure}
  \includegraphics[width=0.9\linewidth]{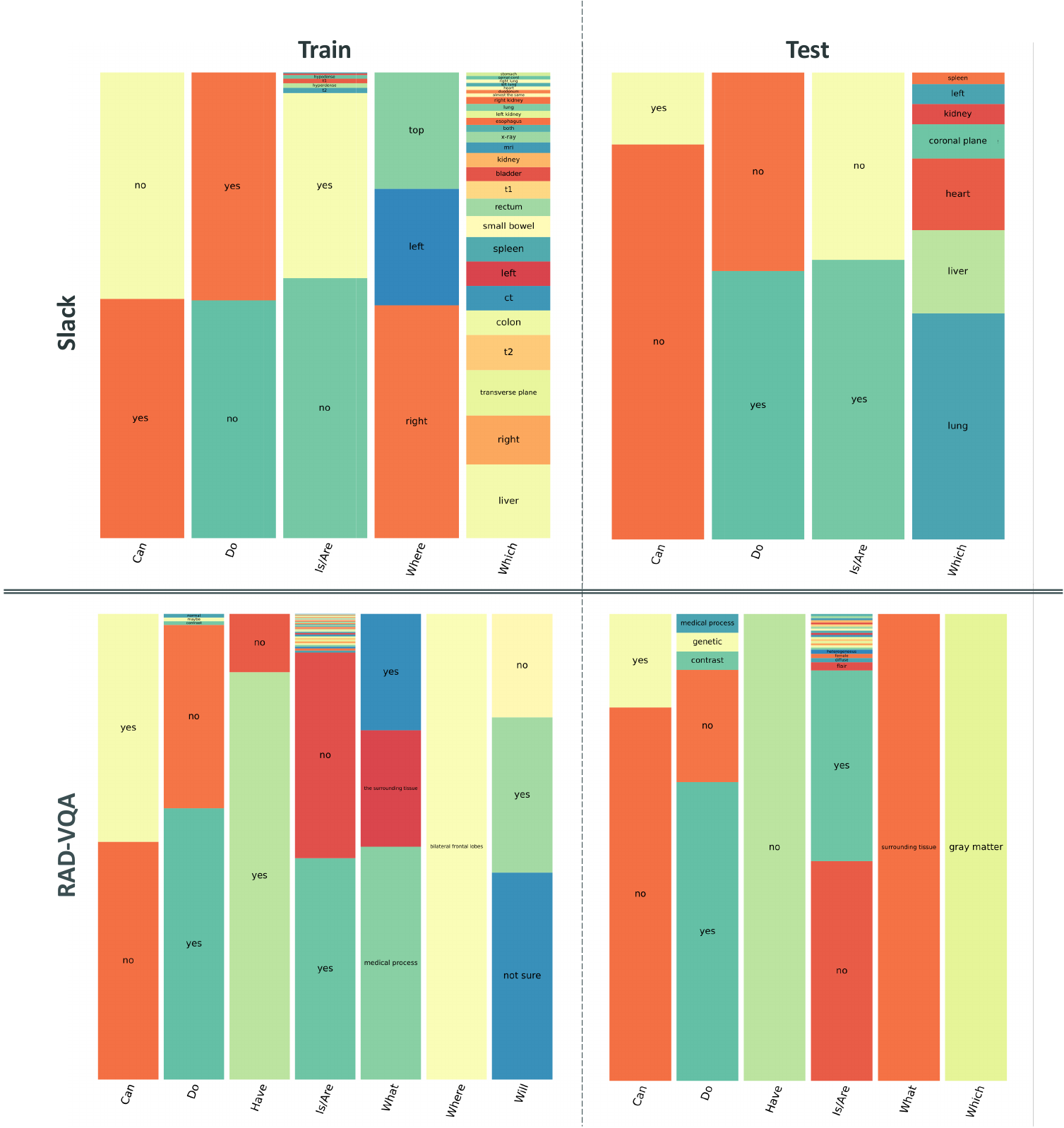}
  \caption{Visualization of data redistribution outcomes. The x-axis denotes the question type, while the bars indicate the proportion of answers corresponding to each question type.}
  \Description{split}
  \label{fig:split}
\end{figure}

Figure ~\ref{fig:split} presents the distribution of answers for each question in the training and testing datasets. A noticeable discrepancy exists between the answers in the training and testing sets for similar types of questions. For instance, within the SLAKE dataset, questions beginning with ``Which" predominantly yield ``liver" as the most frequent answer in the training set, whereas ``lung" is the most frequent answer in the testing set. In the RadVQA dataset, questions starting with ``Have" mostly receive ``yes" as the answer in the training set, while all corresponding answers in the testing set are ``no".

\subsection{Causal Graph Meets Neural Network}

We employ causal reasoning~\cite{causal} to mitigate modality preference bias, where the answer $a$ is predominantly influenced by the question $q$ or the image $v$. The causal representation of this bias is illustrated in Figure~\ref{fig:meet}. Ideally, in MedVQA, both $q$ and $v$ should causally influence the multimodal knowledge $k$, which in turn should determine $a$. However, $a$ is heavily biased by the direct shortcut $q \rightarrow a$, bypassing the image and multimodal knowledge, as shown in Figure~\ref{fig:meet}a. To capture the causal effect of $q$ on $a$, we construct a counterfactual graph representing the scenario where $v$ and $k$ are absent, predicting what $a$ would be under these conditions. Symbols with an asterisk (*) denote the counterfactual scenario. By subtracting the counterfactual graph from the original biased graph, we derive an unbiased causal graph. In this causal graph, $q$ denotes the encoding of the question, $v$ denotes the encoding of the images, and $k$ represents the multimodal fused features. The causal paths $q \rightarrow a$, $v \rightarrow a$, and $k \rightarrow a$ represent the decoders applied to these features to generate the answer. The counterfactual vectors, $q*$ and $v*$, are represented by randomly initialized learnable vectors.

\begin{table*}[t]
\centering
\caption{Performance comparison between MedCFVQA and conventional MedVQA on the SLAKE, RadVQA and SLAKE-CP, RadVQA-CP datasets. The best results are highlighted in bold for each dataset comparison.}
\label{tab:compare}
\begin{tabular}{c|c|c|c|c|c|c|c|c}
    \toprule
     & \multicolumn{4}{c|}{VQA Dataset} & \multicolumn{4}{c}{VQA-CP Dataset} \\
     \cmidrule{2-9}
     Concept & \multicolumn{2}{c|}{SLAKE} & \multicolumn{2}{c|}{RadVQA} & \multicolumn{2}{c|}{SLAKE-CP} & \multicolumn{2}{c}{RadVQA-CP} \\
     \cmidrule{2-9}
     & VQA & CFVQA & VQA & CFVQA & VQA & CFVQA & VQA & CFVQA \\
     \midrule
     Overall & 0.851 & \textbf{0.892} & 0.563 & \textbf{0.574} & 0.337 & \textbf{0.430} & 0.735 & \textbf{0.755} \\
     \midrule
     Abnormal & 0.813 & \textbf{0.861} & 0.463 & \textbf{0.561} & 0.459 & \textbf{0.537} & 0.806 & \textbf{0.900} \\
     Color & 0.250 & \textbf{0.667} & \textbf{1.000} & \textbf{1.000} & \textbf{0.25} & 0.000 & 0.612 & \textbf{0.639} \\
     Modality & \textbf{1.000} & 0.981 & 0.606 & \textbf{0.667} & 0.504 & \textbf{0.697} & 0.250 & \textbf{0.306} \\
     Organ & 0.912 & \textbf{0.915} & \textbf{0.500} & \textbf{0.500} & 0.350 & \textbf{0.458} & 0.500 & \textbf{1.000} \\
     Plane & \textbf{1.000} & \textbf{1.000} & 0.405 & \textbf{0.500} & 0.000 & \textbf{0.028} & 0.556 & \textbf{0.667} \\
     Position & 0.833 & \textbf{0.885} & 0.143 & \textbf{0.286} & 0.000 & \textbf{0.208} & 0.357 & \textbf{0.396} \\
     Size & 0.593 & \textbf{0.750} & 0.535 & \textbf{0.558} & \textbf{0.182} & 0.167 & 0.680 & \textbf{0.794} \\
    \bottomrule    
\end{tabular}
\end{table*}

\begin{figure*}[t]
  \includegraphics[width=0.9\textwidth]{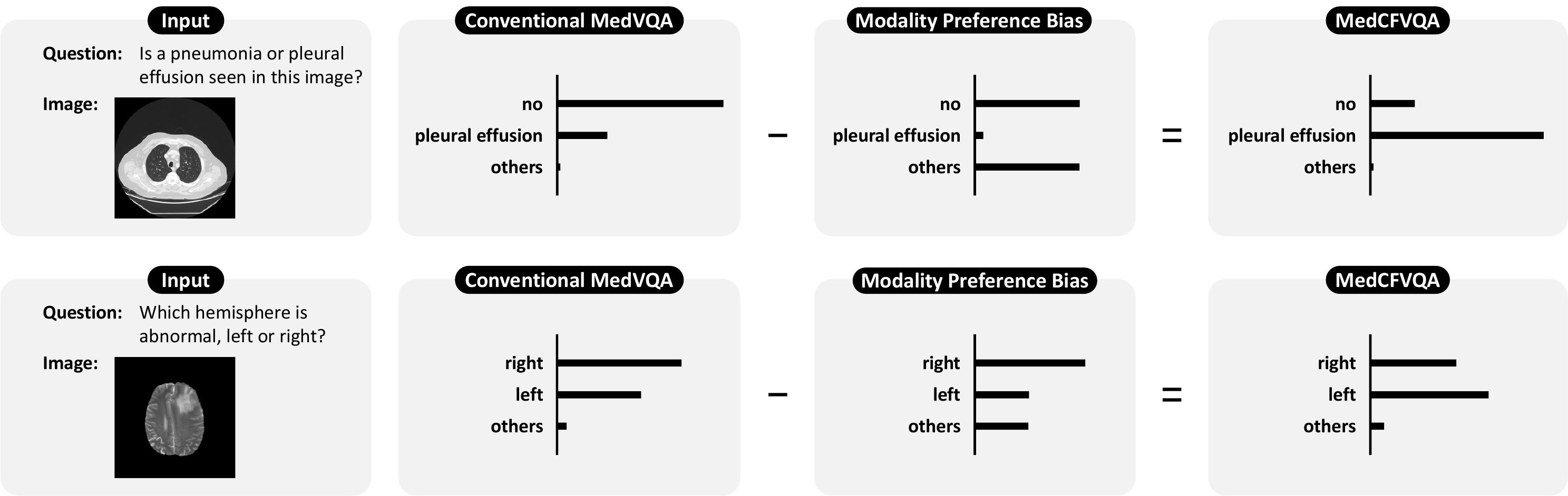}
  \caption{Visualization of the MedCFVQA counterfactual inference process (biased predication - bias = debiased prediction) for mitigating modality preference bias.}
  \Description{MedCFVQA}
  \label{fig:vis}
\end{figure*}

\section{Results and Discussion}

\subsection{Quantitative Analysis}

To validate the efficacy of the proposed method, we conducted a comparative analysis between conventional MedVQA and our MedCFVQA, as detailed in Table~\ref{tab:compare}. The results indicate enhanced performance of MedCFVQA on both SLAKE and RadVQA datasets. Notably, MedCFVQA achieves a significant improvement on the CP versions of these datasets, SLAKE-CP and RadVQA-CP. Specifically, question types such as ``plane" and ``position" in SLAKE-CP, which are unanswerable by conventional VQA, are successfully addressed by MedCFVQA. Conventional VQA methods suffer from modality preference bias, where predictions are predominantly influenced by the direct relationship between questions and answers, disregarding the visual content. Therefore, we observed the performance degradation when they faced totally different questions and answer pairs in training and test sets. Our MedCFVQA removes such bias during inference, which ensures that predictions are grounded in multimodal knowledge, leading to more accurate predictions. Furthermore, the improvements observed in datasets without CP underscore the utility of the counterfactual in enhancing the model's performance in rare cases without compromising its efficacy in more common scenarios.

\subsection{Qualitative Analysis}

To illustrate the effectiveness of the counterfactual debiasing mechanism used in MedCFVQA, Figure 4 presents two examples of the debiasing process. In the first example, the question begins with ``Is" where its answers are typically ``yes" or ``no". However, in this case a multiple-choice question formed to choose between ``pneumonia" or ``pleural effusion". Due to the high modality preference bias, the model is less confident in selecting the correct multiple-choice answer. Through the removal of such bias during counterfactual inference, the model correctly identifies ``pleural effusion" as the appropriate answer. The second example demonstrates a similar improvement by reducing the modality preference bias that assumes most abnormalities occur in the ``right" hemisphere. Through subtracting such bias, the model is able to identify from the corresponding image that the abnormalities are actually in the ``left" hemisphere. Through the examples presented, it can be observed that MedCFVQA effectively eliminates the modality preference bias during inference and ensures predictions are based on multimodal knowledge rather than the shortcut from question to answer.

\section{Conclusion}

We identify modality preference bias in medical visual question answering. To overcome the modality preference bias, we proposed MedCFVQA, which leverages counterfactual reasoning to capture and eliminate bias during inference. Our experiments demonstrate that MedCFVQA outperforms traditional Visual Question Answering (VQA) methods on both the SLAKE, RadVQA, and SLAKE-CP, RadVQA-CP datasets. Additionally, we created MedVQA datasets with changing prior (CP), SLAKE-CP and RadVQA-CP, for modality preference bias benchmarking.

\bibliographystyle{ACM-Reference-Format}
\bibliography{sample-base}










\end{document}